\documentclass[10pt,journal,compsoc]{IEEEtran}

\ifCLASSOPTIONcompsoc
  \usepackage[nocompress]{cite}
\else
  \usepackage{cite}
\fi

\ifCLASSINFOpdf
\else
\fi

\usepackage{color}
\usepackage{bbding}
\usepackage{amsmath}
\usepackage{caption}
\usepackage{amsfonts}
\usepackage{booktabs}
\usepackage{graphicx}
\usepackage{multirow}
\usepackage{algorithm}
\usepackage{subfigure}
\usepackage{algorithmic}

\begin{document}
%
\title{Appearance- and Relation-aware Parallel Graph Attention Fusion Network for Facial Expression Recognition}

\author{Yan Li, Yong Zhao, Xiaohan Xia, and Dongmei Jiang*
\IEEEcompsocitemizethanks{
\IEEEcompsocthanksitem
Y. Li, X. Xia and D. Jiang are with the School of Computer Science, Northwestern Polytechnical University, Beilin District, Xi'an, Shaanxi Province, 710072, China. Y. Li and D. Jiang are also with the Pengcheng Laboratory, Nanshan District, Shenzhen, Guangdong Province, 518055, China. \protect\\
Y. Zhao is with the Zhejiang Lab, Yuhang District, Hangzhou, Zhejiang Province, 311100, China. \protect\\
Corresponding author: Dongmei Jiang
}
}

\markboth{Journal of \LaTeX\ Class Files,~Vol.~14, No.~8, August~2015}%
{Shell \MakeLowercase{\textit{et al.}}: Bare Demo of IEEEtran.cls for Computer Society Journals}

\IEEEtitleabstractindextext{%
\begin{abstract}
The key to facial expression recognition is to learn discriminative spatial-temporal representations that embed facial expression dynamics. Previous studies predominantly rely on pre-trained Convolutional Neural Networks (CNNs) to learn facial appearance representations, overlooking the relationships between facial regions. To address this issue, this paper presents an Appearance- and Relation-aware Parallel Graph attention fusion Network (ARPGNet) to learn mutually enhanced spatial-temporal representations of appearance and relation information. Specifically, we construct a facial region relation graph and leverage the graph attention mechanism to model the relationships between facial regions. The resulting relational representation sequences, along with CNN-based appearance representation sequences, are then fed into a parallel graph attention fusion module for mutual interaction and enhancement. This module simultaneously explores the complementarity between different representation sequences and the temporal dynamics within each sequence. Experimental results on three facial expression recognition datasets demonstrate that the proposed ARPGNet outperforms or is comparable to state-of-the-art methods.
\end{abstract}

}

\maketitle

\IEEEdisplaynontitleabstractindextext

%
\IEEEpeerreviewmaketitle

\IEEEraisesectionheading{\section{Introduction}}
\IEEEPARstart{F}{acial} expression recognition, the task of predicting expressions from facial videos or facial image sequences, has gained significant attention in computer vision and multimedia analysis. Early studies extracted hand-crafted features from the entire face image, such as the Histogram of Oriented Gradients (HOG)~\cite{orrite2009hog} and Local Binary Patterns (LBP)~\cite{zhao2007dynamic}. More contemporary approaches have shifted towards employing Convolutional Neural Networks (CNNs) to learn facial appearance representations~\cite{miao2019deep, bi2022dynamic, li2022tp}, leveraging convolution operations to model local spatial information between neighboring pixels. However, modeling long-range spatial dependencies remains a challenge for CNNs.

Facial expressions are formed by the coordinated movements of muscles in different facial regions~\cite{ekman1978facial}, such as the raising of the eyebrows and the upturning of the corners of the mouth. Therefore, learning high-level relationships between different facial regions is crucial for facial expression recognition. Early studies often manually divided the face into multiple regions, extracted local features from each region, and then fused them through feature concatenation or average pooling~\cite{kim2017isla, zhang2017facial, fan2018multi}. These methods struggled to model the fine-grained interactions between facial regions. Inspired by the success of Vision Transformer in image understanding tasks~\cite{dosovitskiy2020image}, some researchers have started using Transformers to model relationships between different facial regions. These methods~\cite{zhao2021former, xue2022vision, liu2023patch, zhang2024label} typically convert 2D feature maps into 1D feature sequences and then utilize the self-attention mechanism of the Transformer to capture dependencies between regions. However, flattening the 2D feature maps into 1D sequences destroys the intrinsic spatial structure of the face, such as the neighborhood relationships between regions, leading to suboptimal results. Thus, how to effectively learn spatial relationship representations of the face remains a significant challenge.

To further improve facial expression recognition accuracy, researchers have explored complementary information between different facial representations. For instance, some studies fuse global features or confidence scores from different representation sequences to obtain final facial expression prediction~\cite{fan2018video, liu2018multi, pan2024adaptive}. These methods neglect fine-grained interactions between different sequences, limiting their effectiveness. More advanced methods utilize Transformers and attention mechanisms to fuse the complementary information from different representations. For example, a Transformer Encoder with Multimodal Multi-head Attention (TEMMA) is proposed for multimodal continuous emotion prediction~\cite{chen2020transformer}, where Multimodal Multi-head Attention (MMA) is designed to transfer complementary information between different modalities at each moment and Temporal Multi-head Attention (TMA) is proposed to model the temporal dynamics within each modality. Similar architectures are adopted in Multimodal Transformer (MulT)~\cite{tsai2019multimodal} for multimodal sentiment analysis and in Cascade Multi-head Attention (CMHA)~\cite{zheng2022multi} for multimodal emotion recognition. Despite the satisfactory results, these methods treat intra-sequence temporal dynamics and inter-sequence complementarity separately, neglecting direct interactions between representations at different moments across different sequences, resulting in insufficient representations. Therefore, how to effectively fuse the complementary information of different representation sequences remains to be explored.

Recent efforts have also introduced Graph Neural Networks (GNNs)~\cite{velivckovic2018graph} to the task of facial expression recognition. For spatial relation modeling, a study~\cite{jin2021learning} manually crops 20 AU-related Regions of Interest (ROIs) based on facial landmarks and utilizes GNNs to model the relationship between ROIs, which requires pre-extracted landmarks and a complex region division process. FG-AGR~\cite{li2023fg} transforms pixels of 2D facial feature maps into a graph structure via clustering, which disregards facial structure information. For temporal dynamics modeling, the authors of~\cite{xu2024two} learn a sequential graph representation by modeling the relationship from previous to subsequent moments, overlooking the influence of subsequent moments on previous ones. In summary, existing GNN-based approaches for facial expression recognition still lack efficacy in learning facial relational representations and also neglect the complementary information between facial relational and appearance representations.

To address these limitations, we propose an Appearance- and Relation-aware Parallel Graph attention fusion Network (ARPGNet). Firstly, we introduce a face relational graph that encodes facial structure information and utilize graph attention mechanisms to model relationships between different facial regions, resulting in facial relation representation sequences. Secondly, we propose a parallel graph attention fusion module to effectively integrate CNN-based facial appearance representation sequences and GNN-based facial relational representation sequences. By introducing a temporal response scope constraint, this module can simultaneously model intra-sequence dynamics and inter-sequence complementarity, learning compact facial spatial-temporal representations.

In summary, the main contributions of this work are as follows:
\begin{itemize}
\item To complement the CNN-based facial appearance representation, the facial relation representation is learned via graph attention networks to model the relationships among facial regions. By constructing a face region relation graph, valuable face structure information is incorporated into the representation learning process.
\item A parallel graph attention fusion module is designed to simultaneously model the complementary information between appearance and relational representation sequences, as well as temporal dynamics within each sequence, leading to more comprehensive and effective facial spatial-temporal representations.
\item Extensive experimental results on three facial expression recognition datasets demonstrate the effectiveness of the proposed ARPGNet.
\end{itemize}

The remainder of this paper is organized as follows. Section 2 reviews the recent research on facial expression recognition. Section 3 introduces the proposed appearance- and relation-aware parallel graph attention fusion network. The experimental results are analyzed in Section 4, and conclusions are drawn in Section 5.

\section{Related Work}
\subsection{Spatial Representation Learning}
Benefiting from the powerful nonlinear representation ability of CNNs, various pre-trained CNNs have been proposed and utilized for facial expression recognition. For instance, the pre-trained VGG-Face was utilized by Fan et al.~\cite{fan2016video} to extract facial features. In a similar vein, Bargal et al.~\cite{bargal2016emotion} employed a pre-trained ResNet to predict facial expressions, and Savchenko et al.~\cite{savchenko2022classifying} adopted a lightweight EfficientNet to learn facial representations. Some methods have adopted attention mechanisms to focus on important channels and regions of images to learn more effective spatial representations. For example, Liu et al.~\cite{liu2020facial} used ResNet101 with the Convolutional Block Attention Module (CBAM) to extract frame-by-frame features. Li et al.~\cite{li2020attention} designed an attention module to fuse CNN features based on LBP images and RGB images. Recent research has further utilized the potential relationships between emotions to improve performance. For instance, a Multi-Level Dependent Attention Network (MDAN)~\cite{xu2022mdan} was designed to leverage the emotion hierarchy and the correlation between different affective levels and semantic levels, resulting in more robust representations.

While CNNs can effectively learn local appearance information, they face challenges in explicitly modeling the long-range relationships among different facial regions. The combination of muscle movements in different facial regions constitutes the facial expression~\cite{ekman1978facial,li2023towards}. Some studies have divided the whole face into several regions to predict expressions. For example, according to the physical structure of the face, Zhang et al.~\cite{zhang2017facial} divided facial landmarks into four parts (eyebrows, eyes, nose, and mouth) and employed a part-based hierarchical bidirectional recurrent neural network to learn global representations from facial images. Similarly, Fan et al.~\cite{fan2018multi} considered three representative facial regions (eyes, nose, and mouth) and input them into a multi-region ensemble CNN framework, together with the whole face, to learn a global facial representation. Kim et al.~\cite{kim2017isla} evaluated the emotion intensity from the upper and lower facial regions successively. However, existing methods employed either feature concatenation or confidence-based fusion to integrate representations from different facial regions into a global one, which rarely modeled the fine-grained relationship among different facial regions.

Inspired by the success of the Vision Transformer in image understanding tasks~\cite{dosovitskiy2020image,xiao2024towards}, some works have adopted the Transformer to model the relationships between different facial regions. Previous studies~\cite{zhao2021former, liu2023patch} transformed the 2D facial feature maps learned by CNNs into 1D feature sequences and utilized Transformer encoders to model the relationships between different facial regions. Both methods did not take advantage of the structural information inherent in facial images, leading to suboptimal results. Another study~\cite{xue2022vision} incorporated two attentive pooling layers to select the top-K important facial regions for feature extraction in CNN and relationship modeling in Transformer, respectively, while discarding the remaining regions. The hard attention mechanism not only introduces complex hyperparameters but also has the potential to lose valuable clues during the pooling process. Therefore, there is a need for further exploration of facial relationship representation learning methods.

In this study, we divide the face into multiple regions and construct a facial region relation graph based on face structure. A graph attention network is adopted to explicitly model the relationship among different facial regions and learn the face relation representation. Finally, the GNN-based face relation representation and the CNN-based face appearance representation are utilized together for facial expression recognition.

\subsection{Multiple Representation Fusion}
Some studies have adopted multiple facial representation sequences to learn a more comprehensive video-level representation. To conduct a comprehensive comparison, this section reviews the fusion methods for not only facial expression recognition but also multimodal emotion recognition~\cite{li2024aves}. These related research works can be classified into three categories: early fusion, late fusion, and intermediate fusion.

For early fusion, Chen et al.~\cite{chen2015multi} concatenated the embeddings of different modalities at each moment and fed the result into a Long-Short Term Memory (LSTM) network to model the temporal dynamics and obtain a video-level representation. In contrast, late fusion methods are more commonly used due to their practicality~\cite{li2019multimodal}. For instance, Fan et al.~\cite{fan2016video} applied Recurrent Neural Network (RNN) and 3D Convolutional Network (C3D) on a facial image sequence to generate two emotion confidence scores, then used a weighted fusion strategy to produce the final prediction result. Similarly, a weighted fusion strategy was used in~\cite{fan2018video} to integrate the confidence scores from five different CNNs and predict the final emotion category. Wang et al.~\cite{wang2019multi} proposed a multi-attention fusion network that employed intra-modal attention to learn weights for each moment and generate the audio and visual global representations, followed by inter-modal attention to learn weights for both modalities and generate multimodal global features.
However, it is challenging for both early fusion and late fusion to realize fine-grained interactions between different sequences.

Intermediate fusion, on the other hand, enables fine-grained interactions between different representation sequences through a complex network structure~\cite{li2025multimodal,xiao2024oneref}. For example, the authors of~\cite{chen2017multimodal} combined reinforcement learning~\cite{xu2024policy,xu2024time} with the gating mechanism in LSTM to learn more effective multimodal fusion representations. Chen et al.~\cite{chen2020transformer} proposed TEMMA, which includes an MMA layer for multimodal interaction and a TMA layer for temporal dynamic modeling. The interaction between different feature sequences and moments can be realized indirectly by stacking these two layers. MulT~\cite{tsai2019multimodal} and CMHA~\cite{zheng2022multi} employ a cross-modal transformer layer to enable interactions between different modalities, then use a temporal transformer or convolutional layer to model the temporal dynamics within each modality. However, these methods cannot explore the intrinsic interaction between different sequences at different moments as they separate the modeling of inter-modality complementary information and intra-modality temporal dynamics.

This paper proposes a parallel graph attention fusion module, where a parallel fusion graph with a temporal response scope is constructed to allow for simultaneous interactions between different moments and sequences. The complementary information between different sequences and the temporal dynamics within each sequence can interact in the same representation space to obtain more comprehensive spatial-temporal representations.

\section{Appearance- and Relation-aware Parallel Graph Attention Fusion Network}
\begin{figure*}[htbp]
\centering
\includegraphics[width=7in]{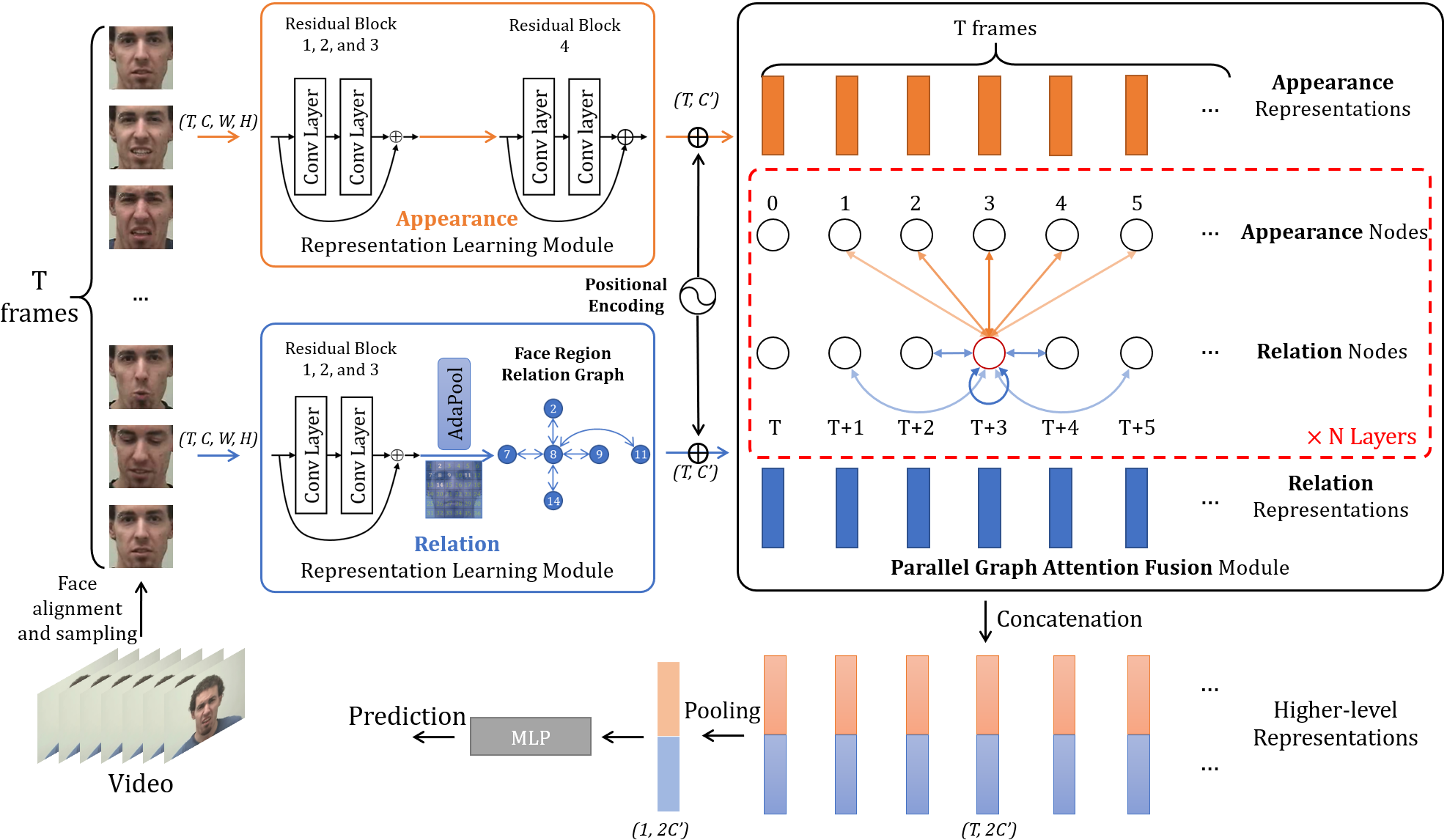}
\caption{The proposed appearance- and relation-aware parallel graph attention fusion network. A facial image sequence is processed by two parallel modules: a CNN-based appearance representation learning module, and a GNN-based relation representation learning module. The former extracts facial appearance information, while the latter models high-level relationships among different facial regions. After introducing temporal position encoding, the two learned facial embedding sequences are then fed into the parallel graph attention fusion module to simultaneously capture the complementary information between the sequences and model the temporal dynamics within each sequence. The mutually enhanced higher-level representations are concatenated and pooled over time to generate a video-level representation, which is then input into an MLP for facial expression recognition.}
\label{fig:framework}
\end{figure*}

\subsection{Overview}
As depicted in Figure~\ref{fig:framework}, the proposed appearance- and relation-aware parallel graph attention fusion network (ARPGNet) for facial expression recognition is comprised of three key components: an Appearance Representation Learning Module (represented by the orange rounded rectangle), a Relation Representation Learning Module (represented by the blue rounded rectangle), and a Parallel Graph Attention Fusion Module (represented by the black rounded rectangle).

The input to ARPGNet is a sequence of face images that have undergone face alignment and frame sampling, with a dimension of $T \times C \times W \times H$. Here, $T$ refers to the number of frames, $W$, $H$, and $C$ indicate the width, height, and number of channels, respectively, of a single face image. The face image sequence is processed by both the convolutional neural network-based Appearance Representation Learning Module to generate a facial appearance embedding sequence ($T \times C^{\prime}$), and the graph neural network-based Relation Representation Learning Module to generate a facial relation embedding sequence ($T \times C^{\prime}$). The value of $C^{\prime}$ represents the dimension of the learned embeddings.

Next, after introducing temporal position encoding, the two facial embedding sequences are fused in the Parallel Graph Attention Fusion Module to mutually enhance the representations by simultaneously considering inter-sequence interactions and intra-sequence temporal dynamics. The resulting higher-level representation sequences are concatenated to form a feature sequence with a dimension of $T \times 2C^{\prime}$. After pooling across the time dimension, a video-level representation ($2C^{\prime}$) is obtained and fed into a multi-layer perceptron (MLP) for facial expression recognition. The details of each module are described in subsequent sub-sections.

\subsection{Appearance Representation Learning Module}
The backbone of an off-the-shelf face recognition network, InsightFace~\cite{deng2018arcface}, is utilized as the Appearance Representation Learning Module. InsightFace\footnote{https://github.com/deepinsight/insightface}, which is based on the ResNet50 architecture and comprises four residual convolutional blocks and two fully-connected layers, is pre-trained on the MS-Celeb-1M face recognition dataset~\cite{guo2016ms}. As a result, it can learn robust facial appearance features. During the training process, the InsightFace backbone is fine-tuned with the guidance of the expression labels, allowing for extracting expression-related appearance information from face images. The output from the first fully connected layer of InsightFace is utilized as the facial appearance representation, resulting in an appearance feature sequence of dimension $T \times C^{\prime}$.

\subsection{Relation Representation Learning Module}
The limited number of training samples in facial expression recognition datasets presents a challenge for learning robust facial relation representations from scratch. To overcome this challenge, in this study, the first three residual convolutional blocks of InsightFace are adopted to extract a facial feature map for each face image. Then, a face relation graph is constructed, and a graph attention mechanism is utilized to model the relationship between different facial regions. It is important to note that the three residual convolutional blocks of InsightFace are also fine-tuned in the training process to learn optimal feature maps for relation representation learning, which allows for more accurate and effective learning of facial relation representations.

\subsubsection{Facial Region Relation Graph Construction}
A general directed graph is defined as $\mathbf{G} = (\mathbf{V}, \mathbf{E})$, where $\mathbf{V}$ denotes a set of nodes and $\left | \mathbf{V} \right |$ is the number of nodes. $\mathbf{E}$ represents a set of edges between connected nodes, which can be represented by an adjacency matrix $\mathbf{A} = [a_{ij}]_{\left | \mathbf{V} \right | \times \left | \mathbf{V} \right |}$.

After the facial feature map is obtained, an adaptive average pooling (AdaPool) layer is utilized to divide it into $P \times P$ patches with equal size. Then, a facial region relation graph $\mathbf{G_{face}}$ with $P^2$ nodes is constructed, where each node corresponds to a single patch of the feature map. Based on the prior knowledge of face structure (e.g., the symmetry of the face~\cite{huang2017beyond}), the edge between \textit{i} and node \textit{j} is defined as:

\begin{equation}
a_{ij}=\left\{
\begin{matrix}
1, & if & i = j; \\
1, & if & patch \ i \ and \ j \ are \ adjacent; \\
1, & if & patch \ i \ and \ j \ are \ symmetrical; \\
0, & else. & \\
\end{matrix}
\right.
\end{equation}

\begin{figure}[htbp]
\centering
\includegraphics[width=3in]{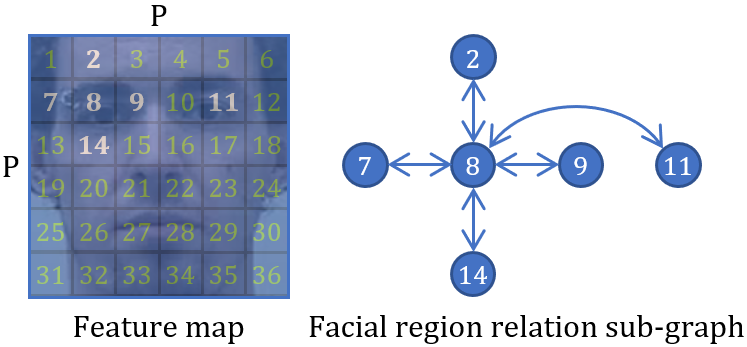}
\caption{Facial feature map after adaptive average pooling layer ($P=6$) and corresponding facial region relation sub-graph of node 8. The connection with itself is omitted for simplicity.}
\label{fig:face-graph}
\end{figure}

Figure~\ref{fig:face-graph} illustrates the facial feature map with 36 patches ($P=6$) and the corresponding facial region relation sub-graph of node 8. As we can see, node 8 is connected to nodes 2, 7, 8, 9, 11, and 14. For simplicity, the connection with itself is omitted.

\subsubsection{Graph Attention Mechanism}
\label{sec:gat}
To explore the relationship between different facial patches, a graph attention mechanism is adopted to adaptively assign weights to neighbors of each node, allowing for the transfer and integration of expression-related information from neighbors to generate discriminative face relation representations.

Given a graph of $N$ nodes, the input to the graph attention layer is a set of node features, denoted as ${\bf x} = \{x_1, x_2, \dots, x_N\}$, where $x_i$ represents the feature of node \textit{i} with dimension $h$. The output of the graph attention layer is a set of higher-level node features, expressed as ${\bf x^{\prime}} = \{x_1^{\prime}, x_2^{\prime}, \dots, x_N^{\prime}\}$, where $x_i^{\prime}$ is the feature of node \textit{i} with dimension $h^{\prime}$. To obtain sufficient expressive power, a shared linear transformation, parameterized by a weight matrix ${\bf W} \in \mathbb{R}^{h^{\prime} \times h}$, is applied to each node. The linear transformation is implemented through a single-layer feedforward neural network without an activation function. Then, the attention coefficient $e_{ij}$ is calculated through a shared attention mechanism, parameterized by a weight vector $\vec{\bf att} \in \mathbb{R}^{2h'}$, to indicate the importance of node \textit{j} to node \textit{i}. The shared attention mechanism is implemented through a single-layer feedforward neural network with the LeakyReLU activation function, which can be expressed as follows:
\begin{equation} 
e_{ij} = \text{LeakyReLU}(\vec{\bf att}^T [{\bf W}x_i\|{\bf W}x_j]),
\end{equation}
where $\cdot^T$ represents the transpose operation, and $\|$ represents the concatenation operation. Therefore, the construction of the facial relationship graph is static, while the weights of the edges in the graph dynamically change based on the input.

To leverage the graph structure information, we inject it into the mechanism by performing masked attention. The attention coefficient $e_{ij}$ is calculated only for node $j \in N_i$, where $N_i$ denotes a set of neighboring nodes of node \textit{i}. The attention coefficients of each node's neighbors are normalized to make them easily comparable across different nodes as follows:
\begin{equation}
\begin{aligned}
\beta_{ij} & = \mathrm{softmax}_i(e_{ij}) \\
& = \frac{\exp\left(\text{LeakyReLU}\left(\vec{\bf att}^T[{\bf W}x_i\|{\bf W}x_j]\right)\right)}{\sum_{k\in\mathcal{N}_i} \exp\left(\text{LeakyReLU}\left(\vec{\bf att}^T[{\bf W}x_i\|{\bf W}x_k]\right)\right)}.
\end{aligned}
\end{equation}
The weighted sum of all neighbor nodes is adopted as the output of each node:
\begin{equation}
x'_i = \sigma\left(\sum_{j\in\mathcal{N}_i} \beta_{ij} {\bf W}x_j\right).
\end{equation}

Moreover, a multi-head attention strategy is adopted to allow the model to jointly attend to information from different representation subspaces and stabilize the learning process of self-attention. Precisely, the single-head attention is executed several times independently, and the features of all heads are averaged to yield the final representation of each node.

To be consistent with the appearance representation learning module, the relation representation learning module utilizes three graph attention layers to model the relationship among different patches. Finally, the relation representations of all patches are averaged to obtain the global relation representation of the facial image, with a dimension of $T \times C^{\prime}$.

\subsection{Parallel Graph Attention Fusion Module}
After obtaining two complementary representation sequences, a Parallel Graph Attention Fusion Module is employed to simultaneously learn the complementary information between different sequences and model the temporal dynamics within each sequence. This module leverages the graph structure and the self-attention mechanism to adaptively focus on salient sequences and moments, thus mutually enhancing the facial representation sequences.

\subsubsection{Positional Encoding}
To enable the model to perceive information about different moments in the sequence, we add positional encodings to the input sequences. Since cosine positional encoding can provide absolute and relative positional information and has better generalization performance, we adopt the original cosine positional encoding in Transformer~\cite{vaswani2017attention}:
\begin{align}
    PE(pos, 2k) &= sin(pos / 10000^{2k/d}) \\
    PE(pos, 2k+1) &= cos(pos / 10000^{2k/d})
\end{align}
where $pos$ represents the position in the sequence, $k$ represents the dimension of the encoding, and $d$ is the size of the encoding dimension. We add the positional encodings to the appearance representation and relation representation sequences before they are fed into the parallel fusion graph.

\subsubsection{Parallel Fusion Graph Construction}
To integrate the facial appearance representations and facial relation representations, a parallel fusion graph $G_{fusion}$ with $2T$ nodes, as depicted in Figure~\ref{fig:framework}, is constructed. The first $T$ nodes correspond to the appearance representations of $T$ frames, and the last $T$ nodes correspond to the relation representations of $T$ frames. Considering that neighboring nodes in the temporal domain exhibit strong relationships, while nodes far apart have limited interaction, the edges between nodes are initialized by a Temporal Response Scope (TRS) as follows:
\begin{equation}
a_{ij}=\left\{
\begin{matrix}
1, & if \ |i\%T-j\%T| \leq TRS;\\
0, & if \ |i\%T-j\%T| > TRS, \\
\end{matrix}
\right.
\end{equation}
where $0 \leq i, j \leq 2T-1$, $\%$ represents the modulo operator, $i\%T$ denotes the temporal moment of node \textit{i}, and $|*|$ represents the temporal distance between nodes \textit{i} and \textit{j}.

\subsubsection{Fusion and Prediction}
The proposed parallel graph attention fusion module achieves mutual enhancement by constructing a unified graph $G_{fusion}$ where nodes from both appearance and relation sequences co-exist and interact. During the feature update process, the set of neighboring nodes $N_i$ for any given node i is not confined to its own sequence. Specifically, as shown in Fig.~\ref{fig:framework}, when updating a relation node at time (T+3), its neighbors can include not only adjacent relation nodes (capturing intra-sequence temporal dynamics) but also temporally-proximate appearance nodes (enabling inter-sequence information exchange). Symmetrically, an appearance node can be enhanced with fine-grained relation details. This direct aggregation of features from the complementary sequence is the core of the mutual enhancement mechanism.

The graph attention mechanism described in the relation representation learning module (Subsection \ref{sec:gat}) is then employed to adaptively assign weights to the neighbors of each node, integrating information from neighbors, and generating two mutually enhanced facial spatial-temporal representation sequences. Specifically, the output feature $\mathbf{x}'_i$ of node $i$ after the fusion is obtained by:
\begin{equation}
    \mathbf{x}'_i = \sigma \left( \sum_{j \in \mathcal{N}_i} \beta_{ij} \mathbf{W} \mathbf{x}_j \right).
    \label{eq:fusion}
\end{equation}
$\mathcal{N}_i$ represents the set of neighboring nodes of node $i$, which contains nodes from both the appearance sequence ($0 \le j < T$) and the relation sequence ($T \le j < 2T$), thus operationalizing the mutual enhancement. $\beta_{ij}$ is the attention weight between node $i$ and $j$, which is calculated according to Equation (3), $\mathbf{x}_j$ denote the appearance or relation features of node $j$. $\mathbf{W}$ is a shared linear transformation or a neural network that maps both the appearance and relation features to the same embedding space.

The attention weights $\beta_{ij}$ are computed using a masked attention mechanism that takes into account the temporal response scope (TRS). The TRS limits the influence of nodes in the temporal domain based on proximity. Therefore, the neighboring nodes in the temporal domain have a strong relationship, while the nodes far apart have a limited impact. Specifically, the attention weights are computed based on the following formula:
\begin{equation}
    \beta_{ij} = \frac{\exp \left(  \mathrm{LeakyReLU}\left(\mathbf{att}^{\top}  [\mathbf{W}\mathbf{x}_i; \mathbf{W}\mathbf{x}_j]\right) \right)}{\sum_{k \in \mathcal{N}_i} \exp \left( \mathrm{LeakyReLU}\left(\mathbf{att}^{\top}  [\mathbf{W}\mathbf{x}_i; \mathbf{W}\mathbf{x}_k]\right) \right)},
\end{equation}
where $\mathbf{att}$ is a shared attention mechanism, which maps the concatenation of node features to a scalar. The attention coefficients $ \beta_{ij} $ can be seen as an importance indicator for information from neighbor node $j$.

After the higher-level representations are concatenated frame by frame and pooled over time, the resulting global representation is fed into an MLP classifier to predict the expression category.

In summary, the mutual enhancement mechanism is achieved by (1) modeling the inter-sequence interactions using the parallel graph structure and adaptive weights, allowing information exchange between the appearance and relation sequences, and (2) modeling the temporal dynamics within each sequence while considering the information from the other sequence. This enables the model to learn more comprehensive and discriminative features for facial expression recognition.

\section{Experiments}
\subsection{Datasets and Evaluation Metrics}
\begin{figure}[htbp]
\centering
\includegraphics[width=3in]{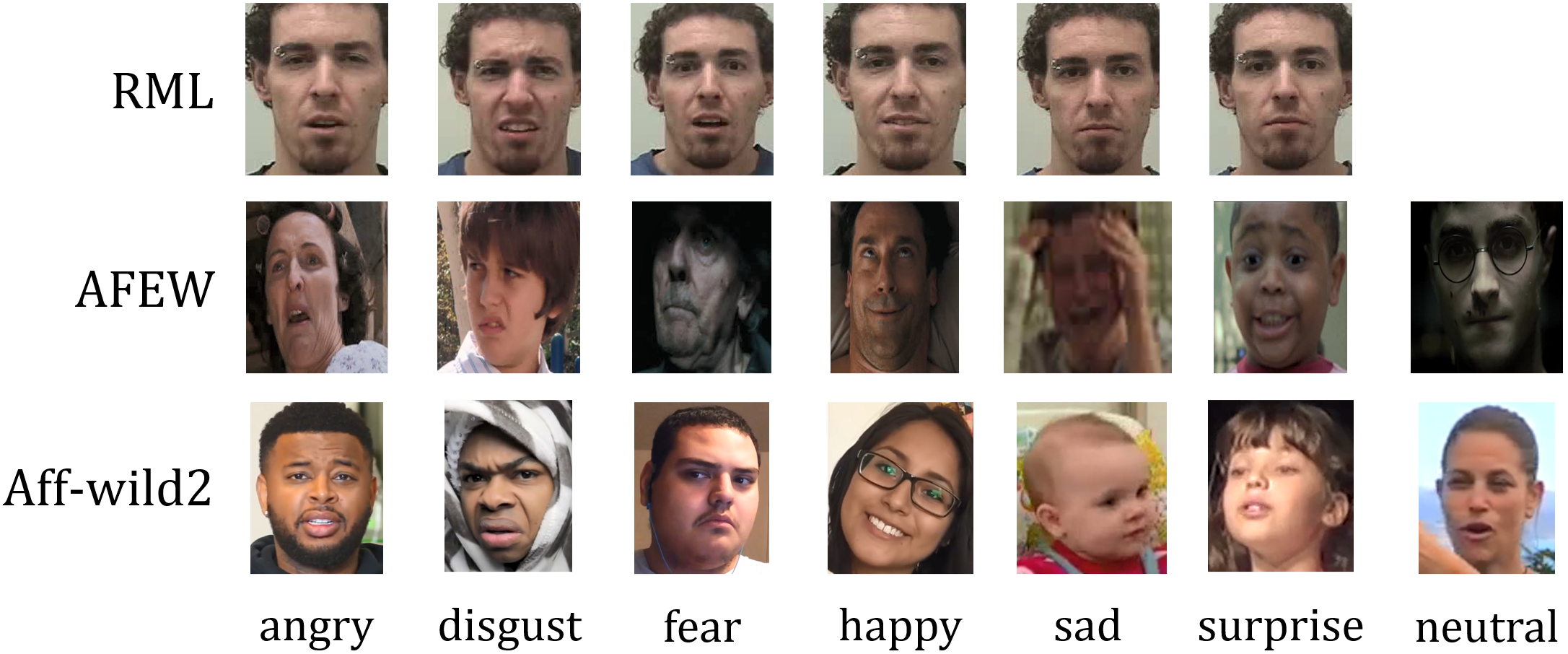}
\caption{Some samples of the three facial expression recognition datasets.}
\label{fig:samples}
\end{figure}

Experiments are conducted on three facial expression recognition datasets to validate the effectiveness of the proposed method, including the in-the-lab RML dataset~\cite{wang2008recognizing}, the in-the-wild AFEW dataset~\cite{dhall2012collecting}, and the in-the-wild Aff-wild2 dataset~\cite{kollias2020analysing}. Figure~\ref{fig:samples} shows some samples of the three datasets.

\subsubsection{RML}
RML is an in-the-lab video-level facial expression recognition dataset collected in the Ryerson Multimedia Research Lab. Eight volunteers from different countries participated in the data collection process. The dataset contains six expressions: anger, disgust, fear, happiness, sadness, and surprise. Each category contains 120 samples with a duration ranging from 3s to 6s. The resolution of the video is $720 \times 480$, and the frame rate is 30.

As suggested in~\cite{zhang2016multimodal, zhang2017learning, ma2019audio, kansizoglou2019active}, the leave-one-subject-out (LOSO) cross-validation protocol is adopted in the experiments and the accuracy is reported. For LOSO cross-validation, the model uses all samples of one subject for evaluation and all samples of the other subjects for training. Due to the dataset containing eight subjects, eight-fold cross-validation is utilized.

\subsubsection{AFEW}
AFEW is an in-the-wild video-level facial expression recognition dataset for the EmotiW challenge. The data was collected from movies and sitcoms with rich variations in lighting, gestures, head movements, and spontaneous expressions~\cite{dhall2019emotiw}. The entire dataset consists of a training set (773 samples), a validation set (383 samples), and a test set (593 samples), with seven expressions: anger, disgust, fear, happiness, sadness, surprise, and neutral. The maximum duration of the video is 7s.

Considering that the annotations on the test set are not publicly available, we train the model, select the hyper-parameters on the training set, and report the results on the validation set following previous work~\cite{zhao2021former, kumar2020noisy, fan2020facial}. Following the official evaluation metric of the EmotiW challenge, accuracy is adopted to evaluate the performance in the experiment.

\subsubsection{Aff-wild2}
Aff-wild2 is an in-the-wild frame-level facial expression recognition dataset for the Affective Behavior Analysis in-the-Wild (ABAW) competition. It claims to be the first dataset annotated for three main behavioral tasks, including valence-arousal assessment, facial action unit detection, and basic expression recognition. The dataset contains 528 samples of 458 individuals collected from YouTube, with a total of 2,813,201 frames. For the basic expression recognition task, the training, validation, and test subsets consist of 51, 11, and 22 videos, respectively. Data partitioning is performed in a subject-independent manner to ensure that a person can appear only in one of the three subsets.

Considering that the annotations on the test set are not publicly available, we train the model, select parameters on the training set and evaluate the results on the validation set. Following the official evaluation metric of the ABAW competition, the result is evaluated as follows:
\begin{equation}
\mathbf{M}=0.33*Acc+0.67*F_1.
\end{equation}

\subsection{Implementation Details}
\subsubsection{Data Processing}
The OpenFace toolkit is utilized to crop, align, and resize the face images to a size of $96 \times 96$. In order to handle the varying lengths of sample sequences in the RML and AFEW datasets, a sparse temporal sampling strategy, as described in~\cite{wang2016temporal}, is adopted. Specifically, 16 face images are randomly sampled from each video during the training process. During the testing phase, 16 frames are uniformly sampled from each video to ensure the consistency of the prediction results. For the Aff-wild2 dataset, a dilated sampling strategy is used following previous work~\cite{kuhnke2020two}. In this approach, eight surrounding frames are sampled from each frame with a dilation rate of three to predict the expression category.

\subsubsection{Training Parameters}
The implementation of the network makes use of the PReLU nonlinear activation function in the residual convolutional blocks, and the LeakyReLU nonlinear activation function, with a negative slope of 0.01, in the graph attention mechanism. The weights of the network are initialized using the Kaiming method. The learning rates for the pre-trained residual convolutional blocks and other parameters are set to 0.0001 and 0.001, respectively. A dropout ratio of 0.25 is employed to prevent overfitting. The optimization process utilizes the Adam optimizer, with beta1 and beta2 set to 0.9 and 0.999, and epsilon set to 1e-08. The loss function used is the cross-entropy loss for RML and AFEW, and the focal loss, following the work in~\cite{thinh2021emotion}, for Aff-wild2 due to its unbalanced expression categories. The network is implemented using the PyTorch deep learning library and runs on NVIDIA TITAN RTX GPUs.

\subsubsection{Implementation of the state-of-the-art fusion models}
\label{sec:SOTA}
We not only compare our method against state-of-the-art methods on each dataset, but also against general multi-representation fusion approaches~\cite{wang2019multi,chen2020transformer,tsai2019multimodal}. To ensure fairness, we utilize our appearance representation learning module and relation representation learning module to learn two facial representations. Then, we use the fusion modules from the literature to integrate the representations and train the entire framework in an end-to-end manner. The results on the three datasets are indicated with an asterisk ($^*$) in Table~\ref{tab:rml}, Table~\ref{tab:afew}, and Table~\ref{tab:affwild2}, respectively.

\subsection{Comparison with State-of-The-Art Methods}
\subsubsection{Results on the RML dataset}
Table~\ref{tab:rml} compares the results of the proposed method with state-of-the-art methods on the RML dataset. Since standard deviations ($\sigma$) were not reported in the original publications for the compared methods, and their source code was unavailable, we report standard deviations only for the methods that we re-implemented.

\begin{table}[htbp]
\centering
\captionsetup{skip=2pt}
\caption{Recognition results on the RML dataset}
\label{tab:rml}
\renewcommand{\arraystretch}{1.1}
\begin{tabular}{c|cc}
\toprule
 & Method & Accuracy$_\sigma$ (\%) \\
\midrule
\multirow{9}{*}{\begin{tabular}[c]{@{}l@{}}Single-\\stream\end{tabular}}
 & LSTM + RL~\cite{kansizoglou2019active} & 60.2 \\
 & Pooling~\cite{zhang2016multimodal} & 60.8 \\
 & Former-DFER~\cite{zhao2021former} & 62.5 \\
 & APViT~\cite{xue2022vision} & 67.6 \\
 & C3D~\cite{zhang2017learning} & 68.1 \\
 & MDAN~\cite{xu2022mdan} & 69.3 \\
 & HSE-NN~\cite{savchenko2022classifying} & 72.2 \\
 & Two-stage~\cite{xu2024two} & 72.5 \\
 & C3D (with audio modality)~\cite{ma2019audio} & 73.9 \\
\midrule
\multirow{8}{*}{\begin{tabular}[c]{@{}l@{}}Multi-\\stream\end{tabular}} & LSTM + LAL~\cite{pan2019deep} & 65.7 \\
 & AM-CNN~\cite{li2020attention} & 68.6 \\
 & PACVT~\cite{liu2023patch} & 70.1 \\
 & Pooling + DBN~\cite{zhang2019learning} & 73.7 \\
 \cmidrule(lr){2-3}
 & Multi-attention~\cite{wang2019multi} & 68.33$_{6.64}$$^*$ \\
 & TEMMA~\cite{chen2020transformer} & 71.11$_{5.41}$$^*$ \\
 & MulT~\cite{tsai2019multimodal} & 73.33$_{4.37}$$^*$ \\
 & ARPGNet (default backbone) & \textbf{76.53}$_{3.05}$$^*$ \\
\bottomrule
\multicolumn{3}{l}{$^*$ means same backbones and different fusion method.}
\end{tabular}
\end{table}

For single-stream facial expression recognition, researchers used pre-trained CNNs to learn appearance representations, and then modeled the temporal dynamics to obtain a video-level representation. For example, in~\cite{zhang2016multimodal} and~\cite{kansizoglou2019active}, a pre-trained CNN was employed to learn frame-level appearance representations, achieving an accuracy of less than 61\%. HSE-NN~\cite{savchenko2022classifying} employed a more robust CNN backbone pre-trained on multiple facial tasks, obtaining an accuracy of 72.2\%. MDAN~\cite{xu2022mdan} incorporated attention mechanisms into CNNs to focus on important feature channels and facial regions, achieving an accuracy of 69.3\%. However, the 2D CNNs used in these studies are unable to capture temporal dynamics while learning spatial representations, resulting in limited recognition performance. Some studies have used pre-trained C3D networks to simultaneously learn local spatial and temporal expression representations. For example, in~\cite{zhang2017learning}, a pre-trained C3D network was applied to face image sequences, achieving an accuracy of 68.1\%. In~\cite{ma2019audio}, an accuracy of 73.9\% was obtained by using a pre-trained C3D network and audio data to reduce cross-modal pollution and redundancy. Nevertheless, these methods do not consider the relationship between different facial regions. Methods based on ViT have achieved limited performance~\cite{zhao2021former, xue2022vision, liu2023patch}, indicating that face structure information is necessary for relation representation learning.

Several works have adopted multi-stream features to enhance expression recognition performance. For instance, in~\cite{pan2019deep} and~\cite{zhang2019learning}, the optical flow modality was introduced to provide more comprehensive expression representations. The global representations from the RGB and optical-flow modalities were fused using a weight-based learnable aggregation layer (LAL)~\cite{pan2019deep} or a deep belief network (DBN)~\cite{zhang2019learning}. One can see that promising facial expression recognition accuracy can be obtained. Meanwhile, AM-CNN~\cite{li2020attention} leveraged attention mechanisms to explore complementary information between RGB and LBP facial images, achieving a performance of 68.6\%.

To further demonstrate the superiority of the proposed parallel graph attention fusion module, we compared it with the state-of-the-art multi-modal fusion methods using the same appearance and relation representation sequences, as described in Section~\ref{sec:SOTA}. Multi-attention~\cite{wang2019multi} achieved an accuracy of 68.33\% by performing inter-modality attention in fusion. TEMMA~\cite{chen2020transformer} improved the accuracy to 71.11\% by first learning the complementary information between modalities at each moment using MMA and then modeling the temporal dynamics of each modality using TMA. MulT~\cite{tsai2019multimodal} further improved the accuracy to 73.33\% by leveraging a cross-modal transformer to explore complementary information and a transformer to model temporal dynamics. The proposed APRGNet outperformed these methods by achieving an accuracy of 76.53\% and a standard deviation of 3.05\%, which was the best among the compared techniques. This is because the proposed method learns inter-sequence complementary information and models intra-sequence temporal dynamics simultaneously.

\subsubsection{Results on the AFEW Dataset}
Table~\ref{tab:afew} presents the results on the AFEW dataset. Previous studies~\cite{vielzeuf2018occam, knyazev2018leveraging, lee2020multi, li2019bi, savchenko2022classifying} used a pre-trained CNN to learn frame-level appearance representations from facial images and then utilized sequential models or statistical methods to obtain video-level representations. MDAN~\cite{xu2022mdan} achieved an accuracy of 54.7\% by leveraging the emotion hierarchy and the correlation between different affective levels and semantic levels. In~\cite{kumar2020noisy}, a three-level attention mechanism was designed to introduce the spatial, temporal, and channel attention into a CNN to focus on important features, and a Noisy Student Training (NST) strategy was introduced to learn more robust representations from unlabeled data, achieving an accuracy of 55.2\%. In~\cite{fan2020facial}, the authors designed a Deeply-Supervised Attention Network (DSAN) that integrated race-, gender-, and age-related information to improve facial expression recognition, achieving an accuracy of 52.7\%. These studies only learned appearance representations with CNNs, which limited their performance in expression recognition. Zhao et al.~\cite{zhao2021former} utilized two transformer layers to model the spatial relationships and temporal dynamics, resulting in an accuracy of 50.9\%. Expression Snippet Transformer (EST)~\cite{liu2021expression} adopted a multi-task learning strategy that combined the shuffled snippet order prediction task with the facial expression recognition task, and used a transformer layer to model the intra-snippet and inter-snippet expression movements, resulting in an accuracy of 54.3\%.

\begin{table}[htbp]
\centering
\captionsetup{skip=2pt}
\caption{Recognition results on the AFEW dataset}
\label{tab:afew}
\renewcommand{\arraystretch}{1.1}
\begin{tabular}{c|cc}
\toprule
 & Method & Accuracy (\%) \\
\midrule
\multirow{12}{*}{\begin{tabular}[c]{@{}l@{}}Single-\\stream\end{tabular}}
 & URNN~\cite{lee2020multi} & 49.0 \\
 & Former-DFER~\cite{zhao2021former} & 50.9 \\
 & Pooling~\cite{vielzeuf2018occam} & 52.2 \\
 & DSAN~\cite{fan2020facial} & 52.7 \\
 & Statistical encoding~\cite{knyazev2018leveraging} & 53.5 \\
 & APViT~\cite{xue2022vision} & 53.8 \\
 & BLSTM~\cite{li2019bi} & 53.9 \\
 & Expression Snippet Transformer~\cite{liu2021expression} & 54.3 \\
 & AEN~\cite{lee2023frame} & 54.6 \\
 & MDAN~\cite{xu2022mdan} & 54.7 \\
 & Two-stage~\cite{xu2024two} & 54.8 \\
 & Three-level attention with NST~\cite{kumar2020noisy} & 55.2 \\
 & MSCM~\cite{li2023multi} & 56.4 \\
 & HSE-NN~\cite{savchenko2022classifying} & 59.3 \\
\midrule
\multirow{10}{*}{\begin{tabular}[c]{@{}l@{}}Multi-\\stream\end{tabular}}
 & Graph-Tran~\cite{zhao2022spatial} & 51.2 \\
 & AM-CNN~\cite{li2020attention} & 51.4 \\
 & RNN + C3D~\cite{fan2016video} & 52.0 \\
 & PACVT~\cite{liu2023patch} & 54.0 \\
 & 4CNNs + LMED~\cite{liu2018multi} & 56.1 \\
 & 5CNNs~\cite{fan2018video} & 57.4 \\
 \cmidrule(lr){2-3}
 & Multi-attention~\cite{wang2019multi} & 53.00$^*$ \\
 & TEMMA~\cite{chen2020transformer} & 54.83$^*$ \\
 & MulT~\cite{tsai2019multimodal} & 55.87$^*$ \\
 & ARPGNet (default backbone) & 57.70$^*$ \\
 & ARPGNet (backbone ablation) & \textbf{60.05} \\
\bottomrule
\multicolumn{3}{l}{$^*$ means same backbones and different fusion method.}
\end{tabular}
\end{table}

For multi-stream expression recognition, some studies have adopted ensemble learning strategies to fuse the prediction results of different networks or features to improve accuracy. Fan et al.~\cite{fan2016video} integrated the results of RNN and C3D using a weighted fusion strategy, achieving an accuracy of 52.0\%. In~\cite{liu2018multi}, the prediction results of four CNNs and the Landmark Euclidean Distance (LMED) were fused, resulting in an accuracy of 56.1\%. Similarly, in~\cite{fan2018video}, the prediction results of five CNNs were fused to achieve an accuracy of 57.4\%. These late fusion methods improved the performance, but they were not capable of implementing simultaneous interactions between different representation sequences at different moments. Moreover, AM-CNN~\cite{li2020attention} utilized spatial-channel attention mechanisms to fuse features from LBP and RGB facial images, obtaining an accuracy of 51.4\%. Compared to other ViT-based methods~\cite{zhao2021former, xue2022vision}, PACVT~\cite{liu2023patch} achieved the best results with an accuracy of 54.0\%, which indicates that appearance features and relation features can complement each other for facial expression recognition.

On the AFEW dataset, the fusion modules of Multi-attention~\cite{wang2019multi}, TEMMA~\cite{chen2020transformer}, or MulT~\cite{tsai2019multimodal} were trained together with the appearance and relation representation learning modules. The results show that the parallel graph attention fusion module obtained the highest accuracy among them, reaching 57.70\%.

For a fair and comprehensive comparison, we further present the result using the same backbone architecture and pre-training datasets. Specifically, HSE-NN~\cite{savchenko2022classifying} employed the EfficientNet-b0~\cite{tan2019efficientnet} architecture and was pre-trained on VGGFace2~\cite{cao2018vggface2} (3M facial images), IMDB-Wiki~\cite{rothe2015dex} (300K facial images), Adience~\cite{eidinger2014age} (15K facial images), UTKFace~\cite{zhang2017age} (23K facial images), AffectNet~\cite{mollahosseini2017affectnet} (1M facial images). With the same backbone, our method achieves state-of-the-art performance, with an accuracy of 60.05\%.

\subsubsection{Results on the Aff-wild2 Dataset}
The experimental results on the Aff-wild2 dataset are presented in Table~\ref{tab:affwild2}. It is worth noting that although the Aff-wild2 dataset includes audio modality and annotations for action units and valence-arousal, we only consider the results obtained using visual modality data and expression annotations for a fair comparison.

\begin{table}[htbp]
\centering
\captionsetup{skip=2pt}
\caption{Recognition results on the Aff-wild2 dataset}
\label{tab:affwild2}
\renewcommand{\arraystretch}{1.1}
\begin{tabular}{c|cc}
\toprule
 & Method & $\mathbf{M}$ \\
\midrule
\multirow{12}{*}{\begin{tabular}[c]{@{}l@{}}Single-\\stream\end{tabular}}
 & MobileNet~\cite{kollias2020analysing} & 0.366 \\
 & CBAM + BLSTM~\cite{liu2020facial} & 0.434 \\
 & Former-DFER~\cite{zhao2021former} & 0.449 \\
 & ResNet (EmotionNet)~\cite{thinh2021emotion} & 0.462 \\
 & SCAN + CCI~\cite{gera2020affect} & 0.465 \\
 & Transformer Encoder~\cite{jin2021multi} & 0.477 \\
 & APViT~\cite{xue2022vision} & 0.494 \\
 & MDAN~\cite{xu2022mdan} & 0.510 \\
 & R(2+1)D~\cite{kuhnke2020two} & 0.515 \\
& Two-stage~\cite{xu2024two} & 0.517 \\
 & HSE-NN~\cite{savchenko2022classifying} & 0.521 \\
 & Deviation Learning Network~\cite{zhang2021prior} & \textbf{0.677} \\
\midrule
\multirow{7}{*}{\begin{tabular}[c]{@{}l@{}}Multi-\\stream\end{tabular}}
 & AM-CNN~\cite{li2020attention} & 0.459 \\
 & PACVT~\cite{liu2023patch} & 0.517 \\
 \cmidrule(lr){2-3}
 & Multi-attention~\cite{wang2019multi} & 0.508$^*$ \\
 & TEMMA~\cite{chen2020transformer} & 0.531$^*$ \\
 & MulT~\cite{tsai2019multimodal} & 0.536$^*$ \\ 
 & ARPGNet (default backbone) & 0.547$^*$ \\
 & ARPGNet (backbone ablation) & 0.628 \\
\bottomrule
\multicolumn{3}{l}{$\mathbf{M}=0.33*Acc+0.67*F_1$} \\
\multicolumn{3}{l}{$^*$ means same backbones and different fusion method.}
\end{tabular}
\end{table}

In~\cite{kollias2020analysing}, the authors utilized a MobileNet architecture to learn frame-level face embeddings and predict the expression categories, achieving a result of 0.366 without considering the temporal dynamic information. In~\cite{liu2020facial}, the authors incorporated the convolutional block attention module (CBAM) into ResNet101 to focus on critical regions and channels and employed a bidirectional LSTM (BLSTM) to model the temporal dynamics, resulting in an M of 0.434. The authors of~\cite{gera2020affect} introduced both spatial and channel attentions and formulated a complementary context information (CCI) branch to enhance the discrimination ability of learned representations, resulting in a final M of 0.465. Similarly, MDAN~\cite{xu2022mdan} achieved a result of 0.510. AM-CNN~\cite{li2020attention} acquired a relatively limited performance of 0.459, indicating that LBP features are not robust enough for in-the-wild facial expression recognition. In~\cite{jin2021multi}, the authors utilized an IResNet100 architecture to learn the facial expression representations and employed a transformer encoding layer to capture the temporal dynamics, achieving a result of 0.477. The authors of~\cite{kuhnke2020two} designed an R(2+1)D network to model the spatial and temporal dynamics in face image sequences, achieving an M of 0.515. However, the performance of these CNN-based methods is limited due to their difficulty in exploring the relationships among different facial regions and learning facial relation representations.

Some studies fine-tuned pre-trained models using additional emotion datasets to learn more emotion-related representations. In~\cite{thinh2021emotion}, the authors fine-tuned a ResNet50 architecture pre-trained on the EmotionNet dataset to learn facial embeddings and predict expressions, resulting in an M of 0.462. HSE-NN~\cite{savchenko2022classifying} was finetuned on the AffectNet dataset, acquiring a result of 0.521. In~\cite{zhang2021prior}, the authors developed a deviation learning network (DLN) to generate a compact and continuous expression embedding space disentangled from the identity factor. The DLN was pre-trained on several emotion recognition, valence-arousal prediction, and AU prediction datasets, resulting in the highest M of 0.677. Our proposed ARPGNet, without additional emotion datasets for training, achieved a result of 0.547, outperforming the state-of-the-art fusion methods in Multi-attention, TEMMA, and MulT.

For a fair and comprehensive comparison, we further present the result using the same backbone architecture. Specifically, DLN employed the Inception-Resnet architecture and was pre-trained on FaceNet~\cite{schroff2015facenet} (100-200M facial images), FECNet~\cite{vemulapalli2019compact} (500K facial image triplets), BP4D~\cite{zhang2014bp4d}, BP4D+~\cite{zhang2016multimodal}, DFEW~\cite{jiang2020dfew} (16K video clips), and AffectNet~\cite{mollahosseini2017affectnet} (1M facial images). Due to the unavailability of the pre-trained models in~\cite{zhang2021prior} and the inaccessibility of the FECNet, BP4D, and BP4D+ datasets, we only pre-trained the Inception-Resnet network on FaceNet, DFEW, and AffectNet datasets, improving the M value from 0.547 to 0.628.  Although our method did not achieve the best result, the backbone ablation result demonstrates its potential.

\subsection{Ablation Study}
\subsubsection{Impact of Module Removal}
We present the results of the module ablation experiments on the AFEW dataset in Table~\ref{tab:ablation}. To begin, we perform single-stream facial expression recognition experiments using either the appearance representation sequence (from the appearance module) or the relation representation sequence (from the relation module), with the mean pooling being employed to model the temporal dynamics within each sequence. The results indicate that the appearance representations achieve higher accuracy than the relation representations, likely since the appearance representation learning module is initialized by InsightFace (pre-trained on a large face image dataset) and fine-tuned on the AFEW dataset, whereas the relation representation learning module is trained from scratch on the AFEW dataset.

\begin{table}[htbp]
\centering
\captionsetup{skip=2pt}
\caption{Module ablation on the AFEW dataset}
\label{tab:ablation}
\renewcommand{\arraystretch}{1.1}
\begin{tabular}{ccccc}
\toprule
  \begin{tabular}[c]{@{}c@{}}Appearance\\ Module\end{tabular} &
  \begin{tabular}[c]{@{}c@{}}Relation\\ Module\end{tabular} &
  \begin{tabular}[c]{@{}c@{}}Fusion\\ Module\end{tabular} &
  \begin{tabular}[c]{@{}c@{}}Temporal\\ Scope\end{tabular} &
  Acc (\%) \\
\midrule
\checkmark & - & - & - & 50.65 \\
- & \checkmark & - & - & 47.52 \\
\checkmark & \checkmark & - & - & 51.96 \\
\checkmark & \checkmark & \checkmark & - & 55.35 \\
\checkmark & \checkmark & \checkmark & \checkmark & 57.70 \\
\midrule
\end{tabular}
\end{table}

For the baseline multi-stream facial expression recognition experiments, we concatenate the appearance and relation representations frame by frame, pooling the results over time as the input to an MLP. This approach results in an accuracy of 51.96\% and a 1.31\% improvement, which indicates the complementary information between the appearance feature and the relation feature. By incorporating the proposed parallel graph attention fusion module, the accuracy is improved to 55.35\%, even without the constraint of temporal response scope, demonstrating the effectiveness of the fusion module in learning the intra-sequence temporal dynamic information and the inter-sequence complementary information. When the TRS is added to limit the scope of interaction among the nodes, the accuracy further increases to 57.70\%, indicating that TRS effectively limits the transfer of distant noise and leads to improved embeddings for facial expression recognition.

\subsubsection{Impact of Different Patch Counts}
Given that the feature map size before constructing the facial relationship graph is 12×12, we divide it into 6×6 (36) patches by default, with each patch measuring 2×2. To further evaluate the impact of different patch counts on experimental results, we present the findings on the AFEW dataset in Table~\ref{tab:ablation1}. The results indicate that both excessively small and overly large patches lead to lower facial expression recognition accuracy.

\begin{table}[htbp]
\centering
\captionsetup{skip=2pt}
\caption{Patch count ablation on the AFEW dataset}
\label{tab:ablation1}
\renewcommand{\arraystretch}{1.1}
\begin{tabular}{ccc}
\toprule
Patch Count & Patch Size & Acc (\%) \\
\midrule
12x12 (144) & 1x1 & 57.18 \\
6x6 (36) & 2x2 & \textbf{57.70} \\
4x4 (16) & 3x3 & 56.91 \\
3x3 (9) & 4x4 & 55.61 \\
\bottomrule
\end{tabular}
\end{table}

The underlying reason is that when the patch size is too small (e.g., 1×1), there are a large number of nodes in the face relationship graph, and each patch contains only very limited information, making it difficult to further learn meaningful relational features. Conversely, when the patch size is too large (e.g., 4×4), the graph network struggles to capture fine-grained facial relationships effectively. Consequently, an optimal patch size strikes a balance between these extremes, yielding the best performance.

\subsubsection{Impact of Different Facial Relation Graphs}
We compare the proposed graph construction method with some AU-based graph construction approaches, and Table~\ref{tab:ablation2} presents the results on the AFEW dataset. Specifically, RA-UWML~\cite{chen2021region} defines three Regions of Interest (ROIs) on the face, whereas the other two methods~\cite{jin2021learning,chang2022knowledge} utilize facial landmark coordinates to extract finer-grained patches. As we can see, our method achieves the highest accuracy. Moreover, the proposed graph construction method does not rely on pre-extracted facial landmark coordinates, making it simple and efficient.

\begin{table}[htbp]
\centering
\captionsetup{skip=2pt}
\caption{Facial relation graph ablation on the AFEW dataset}
\label{tab:ablation2}
\renewcommand{\arraystretch}{1.1}
\begin{tabular}{ccc}
\toprule
Method & Landmark & Acc (\%) \\
\midrule
RA-UWML~\cite{chen2021region} & & 52.22 \\
DDRGCN~\cite{jin2021learning} & \checkmark & 54.31 \\
Chang et al.~\cite{chang2022knowledge} & \checkmark & 55.61 \\
Our & & \textbf{57.70} \\
\bottomrule
\end{tabular}
\end{table}

\subsection{Inference Time Analysis}
\begin{figure}[htbp]
\centering
    \includegraphics[width=3.5in]{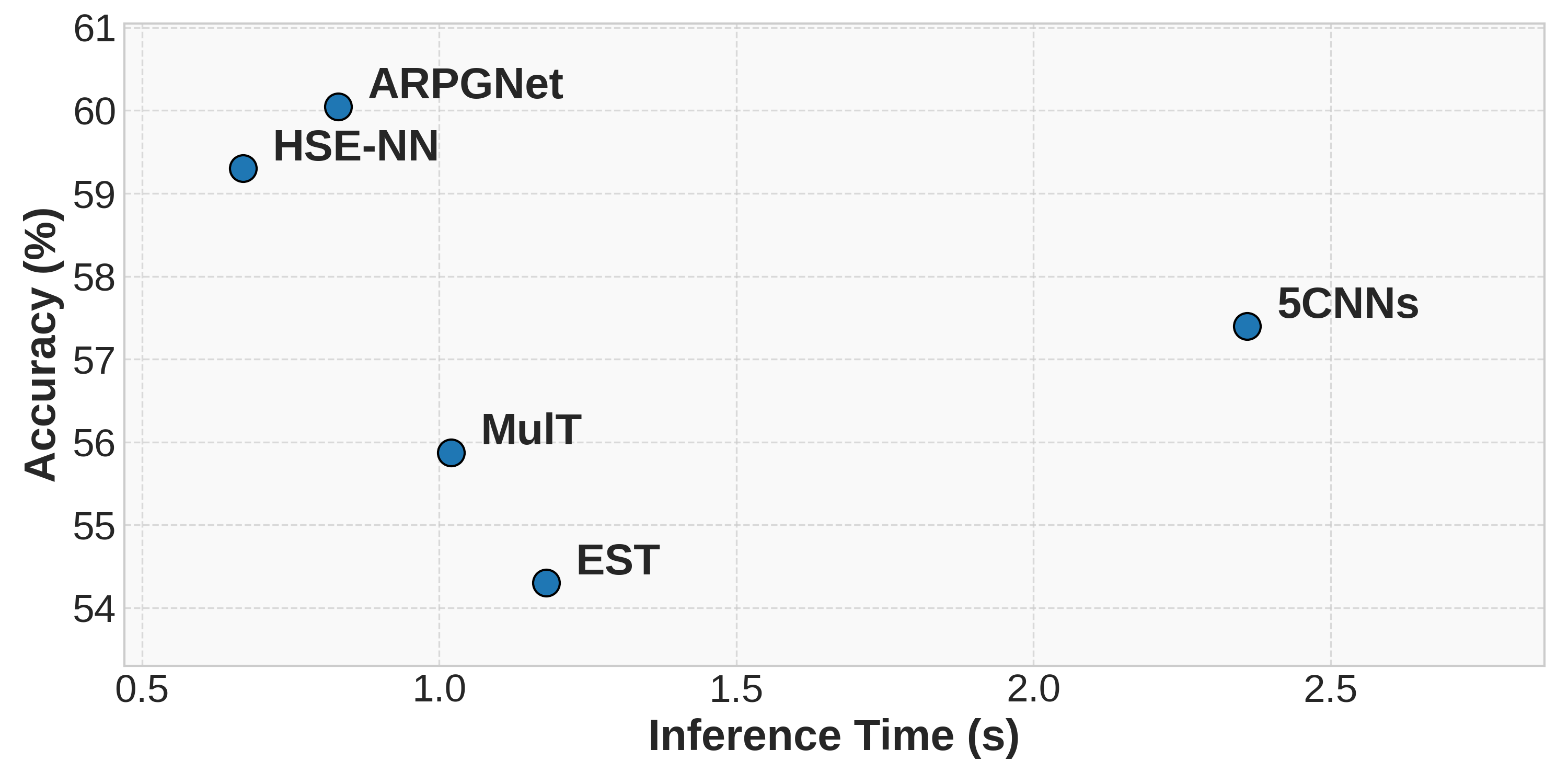}
\caption{Comparison of model accuracy and inference time.}
\label{fig:inference}
\end{figure}

To further evaluate the efficiency of different approaches, we compared their inference time on the AFEW dataset. Since most of the methods do not provide publicly available implementations, we re-implemented several representative models~\cite{fan2018video,wang2019multi,liu2021expression,savchenko2022classifying} and measured their inference time under the same experimental settings. Each method was executed three times, and the average inference time was recorded. The results are illustrated in Fig.~\ref{fig:inference}.

In the figure, methods located in the upper-left region represent favorable trade-offs, achieving high accuracy with low inference time, while those in the lower-right corner indicate less efficient performance. As shown, our proposed method demonstrates the highest accuracy while maintaining a highly competitive inference time. In contrast, the 5CNNs approach~\cite{fan2018video} shows the longest inference time with only moderate accuracy. Moreover, the quadratic complexity associated with the self-attention mechanism limits the inference speed of our proposed method. Consequently, a direction for future research is to investigate alternative model architectures, particularly those with linear complexity~\cite{li2025alignmamba}, aiming for improved computational efficiency.

\subsection{Visualization}
We conduct visualization analysis from three aspects: temporal attention, spatial attention, and the final prediction results.

\begin{figure*}[htbp]
\centering
\subfigure[Visualization of attention scores from the neighbors to the target node on frame 10 of the relation sequence. Frames 7 to 10 of the appearance representation, as well as frames 8 and 10 of the relation representation, contribute more to the target node.]{
    \includegraphics[width=6in]{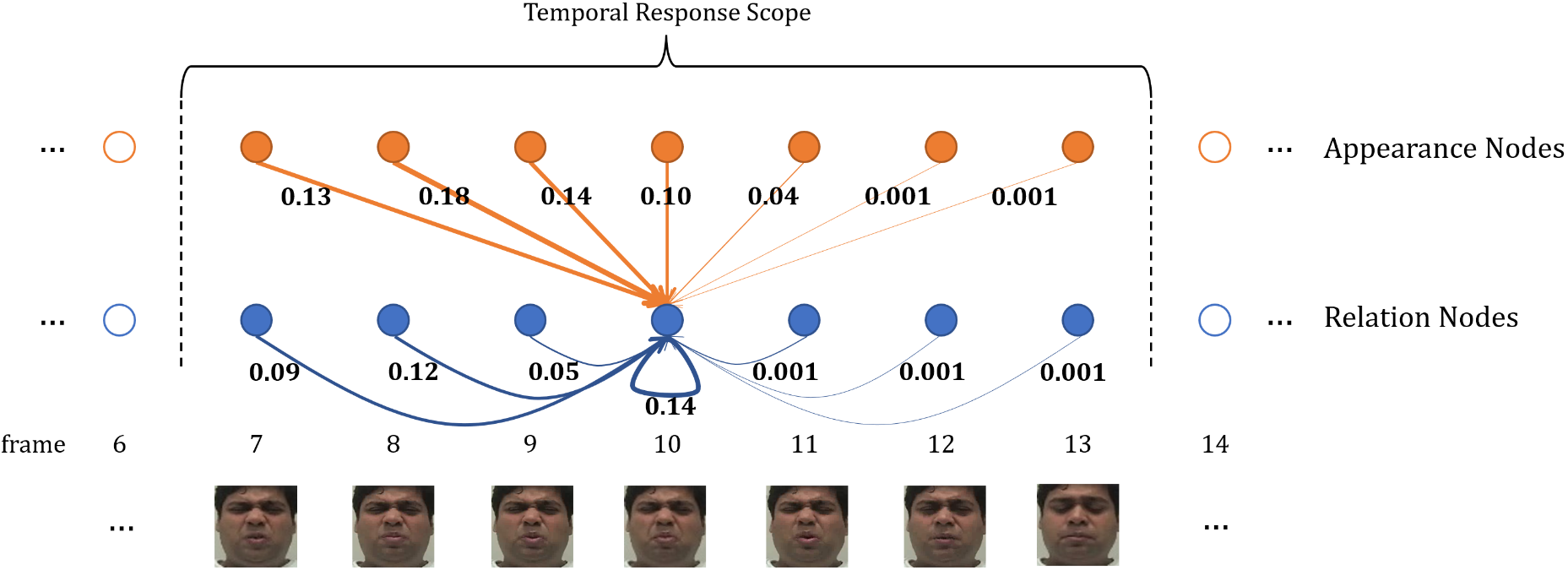}
}
\\
\subfigure[Visualization of attention scores from the target node on frame 10 of the relation sequence to its neighbors. It contributes more to the nodes of the appearance and relation representations behind its moment.]{
    \includegraphics[width=6in]{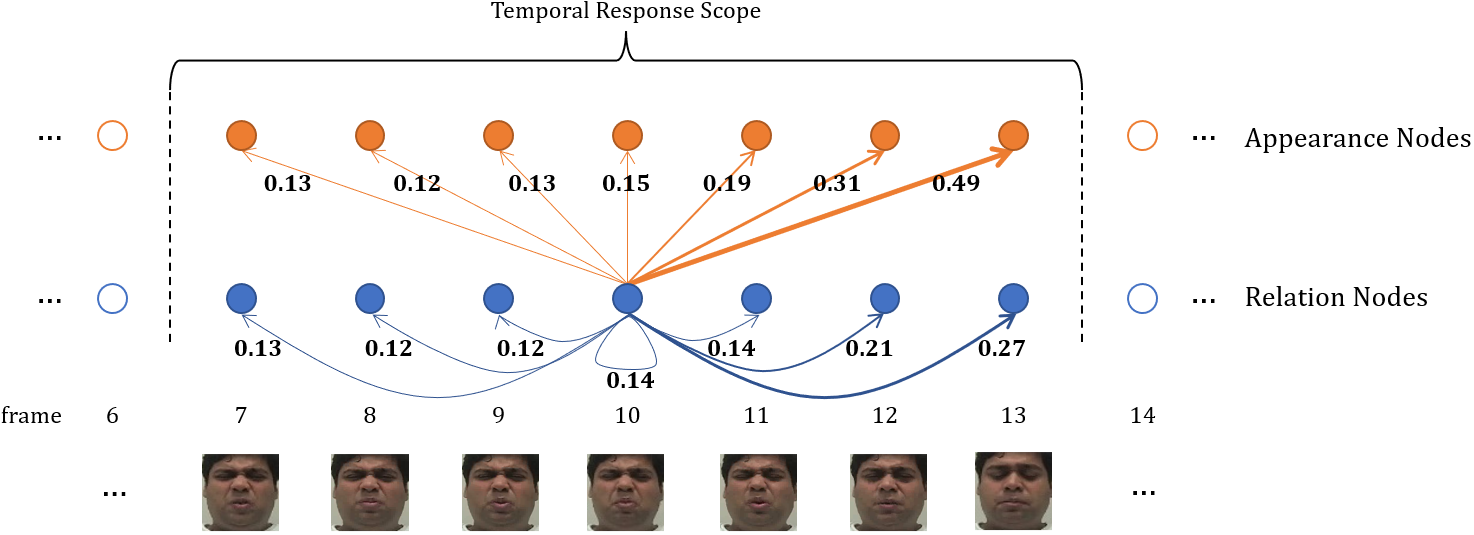}
}
\caption{Visualization of attention scores between the relation representation node at frame 10 (marked as the target node) and its neighbors. It can be seen that inter-sequence complementary information and intra-sequence temporal dynamics can help each other in representation learning.}
\label{fig:vis12}
\end{figure*}

\subsubsection{Temporal Attention}
Figure~\ref{fig:vis12} illustrates an example of the attention scores between a target node (frame 10 of the relation sequence) and its neighboring nodes. Figure~\ref{fig:vis12} (a) displays the attention scores from the neighbors within the temporal response scope to the target node. It is evident that more attention is allocated to the frames preceding the target node as compared to those following the target node. This is expected as the target node primarily learns the emotional representation from historical information. Additionally, the attention scores from the appearance representations of frames 7, 8, and 9 are relatively high, suggesting that the relation representation of the target node at frame 10 gains much complementary information from the other sequence. Figure~\ref{fig:vis12} (b) presents the attention scores from the target node at frame 10 of the relation representation to its neighbors within the temporal response scope. It is noticeable that the target node has a more significant impact on the following frames than the preceding frames. Furthermore, the attention scores from the target node to the appearance representations at frames 11 to 13 are high, indicating that these appearance representations receive much supplementary information from the relation representation of the target node. In conclusion, from Figure~\ref{fig:vis12}, it can be inferred that the proposed parallel graph attention fusion module effectively promotes the learning of complementary information from the other sequence.

\begin{figure}[htbp]
\centering
\includegraphics[width=3.5in]{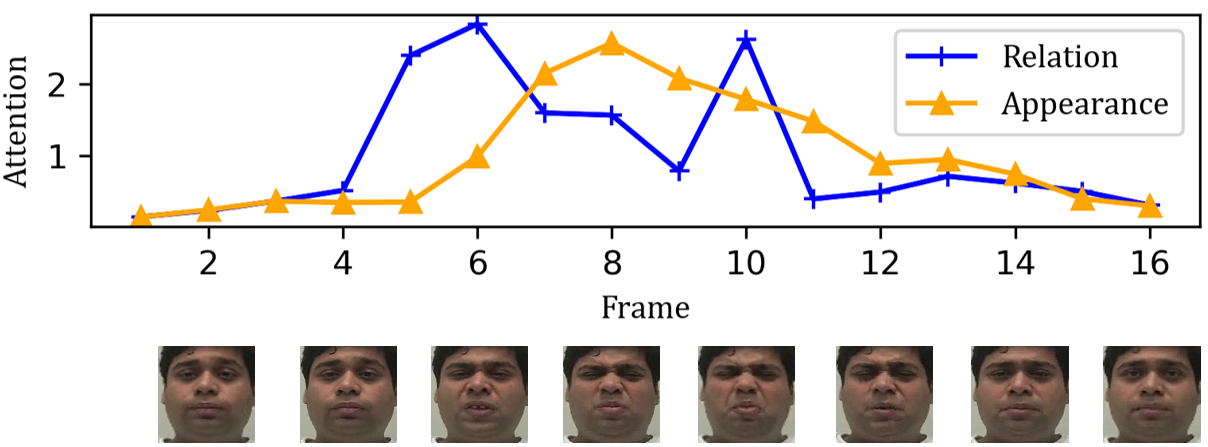}
\caption{Visualization of attention scores on the appearance representation nodes and relation representation nodes in the parallel graph fusion module.}
\label{fig:vis}
\end{figure}

To show the attention weights over time, for each node, we sum up the attention scores on the edges from this node to all its neighboring frames within the temporal response scope. The results are depicted in Figure~\ref{fig:vis}, where the orange line represents the results of the appearance nodes, and the blue line represents those of the relation nodes, respectively. It can be seen that the proposed parallel graph attention fusion module is capable of automatically allocating more attention to the expression-related salient frames in the middle for both appearance representations and relation representations.

\begin{figure*}[htbp]
\centering
\subfigure[Visualization of attention scores from the neighbors to the region 8.]{
    \includegraphics[width=6in]{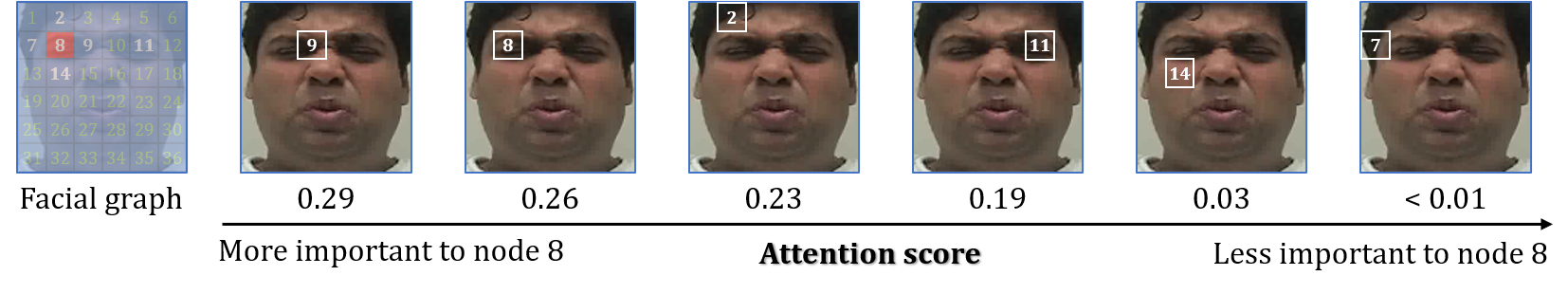}
}
\\
\subfigure[Visualization of attention scores from the region 8 to its neighbors.]{
    \includegraphics[width=6in]{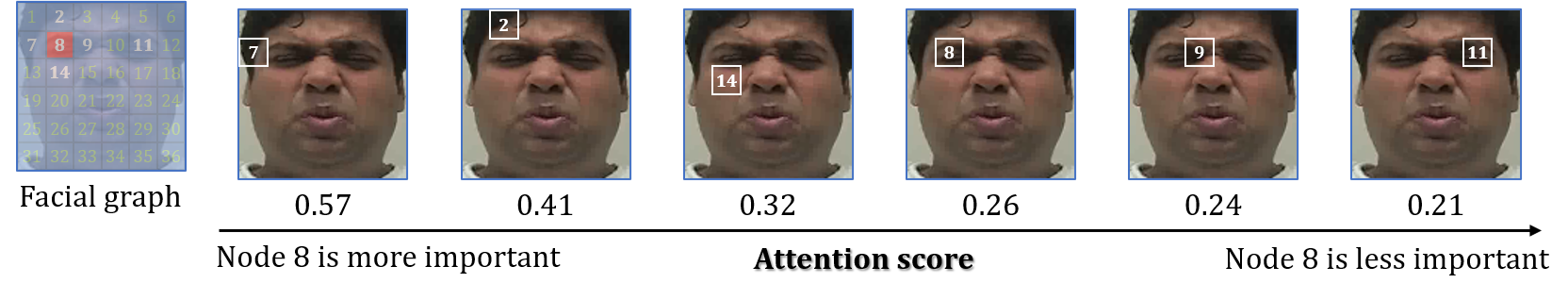}
}
\caption{Visualization of attention scores between the region 8 and its neighbors.}
\label{fig:vis34}
\end{figure*}

\subsubsection{Spatial Attention}
We also demonstrate the learned spatial attentions at the 10th frame. Taking the facial region 8 as an example, Figure~\ref{fig:vis34} (a) shows the importance of neighboring regions to region 8. It can be observed that region 9 is the most important to region 8, with an attention score of 0.29, because this region contains wrinkles on the inner eyebrows and nose bridge. Region 7 is the least important to region 8, with an attention score of less than 0.01, because it contains little facial expression information. Figure~\ref{fig:vis34} (b) shows the importance of region 8 to its neighboring regions. It can be seen that region 8 is most important to region 7 compared to other regions. This is because, for the neighboring regions of region 7 (i.e., regions 1, 7, 8, 12, and 13), region 8 contains the most expression-related information and thus receives more attention.

\subsubsection{Prediction Analysis}
\begin{figure*}[htbp]
\centering
\includegraphics[width=6in]{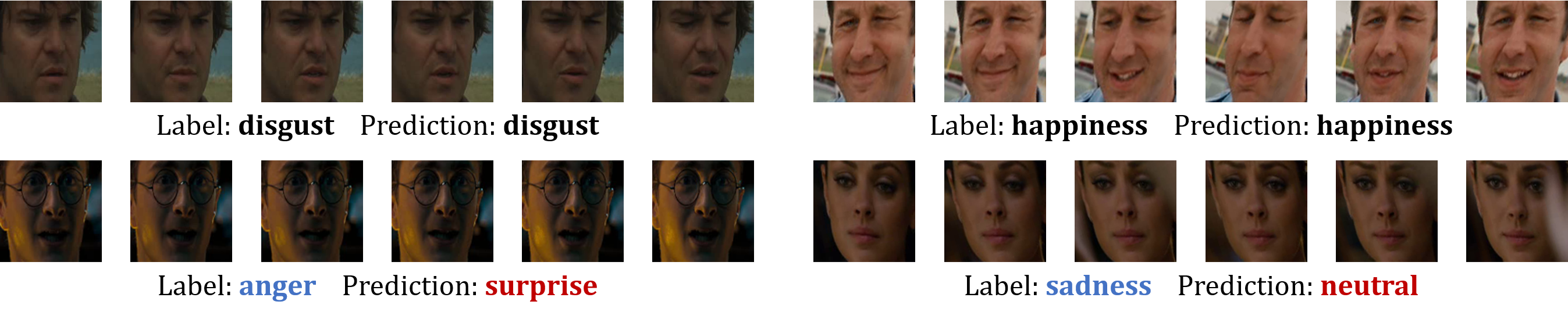}
\caption{Two samples that are correctly classified and two samples that are misclassified.}
\label{fig:pred}
\end{figure*}

We present prediction results for more samples. Figure~\ref{fig:pred} (top) shows two correctly classified samples, corresponding to disgust and happiness. Figure~\ref{fig:pred} (down) shows two misclassified samples. One classified anger as surprise, which may be due to poor lighting that makes it difficult to distinguish the facial expression. The other classified sadness as neutral. From the corresponding facial image sequences, it can be seen that the expression of sadness is not pronounced and is limited to the movement of the eyebrows, which results in an incorrect facial expression recognition result. In conclusion, the proposed method still has certain limitations when dealing with some extreme or difficult samples.

\subsection{Discussion on Temporal Modeling}
Our choice of a temporally-scoped graph over conventional architectures like LSTMs or standard Transformers is motivated by efficiency and task suitability. Standard Transformers, with their global attention, incur significant computational overhead for long sequences. More importantly, forcing the model to attend to distant, potentially irrelevant frames can introduce noise and degrade performance, a phenomenon observed in our experiments (Section~\ref{sec:SOTA}). The dynamics of facial expressions are predominantly local; thus, our approach of using a Temporal Response Scope (TRS) focuses computational resources on the most informative short-term temporal patterns. Compared to recurrent models like LSTMs, which are limited by sequential processing and potential gradient issues, our graph-based method allows for parallel computation and captures bidirectional relationships within a local window. This design strikes an effective balance between modeling crucial temporal dynamics and maintaining computational efficiency, which contributes significantly to the superior performance of our proposed ARPGNet.

\section{Conclusion and Future Work}
\subsection{Conclusion}
In this paper, we propose an appearance- and relation-aware parallel graph attention fusion network for the task of facial expression recognition. To complement the appearance representation learned by CNNs, we construct a facial region relation graph based on facial patches and use graph attention mechanisms to learn the relation representations from these patches. Additionally, to integrate the appearance representation sequence and the relation representation sequences, we introduce a parallel graph attention fusion module, which mutually enhances the representations by modeling both inter-sequence interactions and intra-sequence temporal dynamics. Our extensive experiments on three facial expression recognition datasets demonstrate the superiority of our proposed ARPGNet compared to state-of-the-art methods.

\subsection{Limitations and Future Work}
Despite the promising results achieved by the proposed ARPGNet framework in integrating appearance and relation representations for facial expression recognition, several limitations warrant discussion and suggest avenues for future research.

Firstly, the robustness of the current model under extreme conditions, such as severe occlusions (e.g., face masks, hands covering parts of the face) or highly challenging illumination variations (e.g., very low light, strong backlighting), might be limited. While the graph attention mechanism aids in modeling relationships between visible regions, significant information loss due to occlusion or distortion due to lighting can potentially degrade the quality of both the appearance features extracted by the CNN backbone and the learned relation representations. To mitigate this, future work could explore the integration of more advanced representation learning techniques, such as Masked Auto-Encoders (MAEs)~\cite{he2022masked} or Multimodal Large Models (MLMs)~\cite{li2025multimodal}. These methods have shown potential in learning holistic and robust representations even from partially visible data, which could enhance the model's resilience to occlusions and adverse conditions.

Secondly, the generalization capability of the model across significantly different domains remains a challenge. As often observed, models trained primarily on datasets from one domain (e.g., controlled in-the-lab videos) may experience a performance drop when directly applied to another substantially different domain (e.g., unconstrained in-the-wild videos) due to domain shift. While our experiments cover both types of datasets, improving the inherent cross-domain generalization is a crucial aspect for practical applications. To address this limitation, investigating unsupervised or self-supervised domain adaptation techniques~\cite{li2024prompt,xu2024novel}, particularly Test-Time Adaptation (TTA) strategies, could be a fruitful direction. TTA methods aim to adapt the pre-trained model to the characteristics of the unlabeled target test data during inference, potentially improving performance in unseen domains without requiring retraining or target domain labels.

We consider these aspects as important directions for future investigation to further enhance the robustness and applicability of our facial expression recognition framework in diverse real-world scenarios.


%



\ifCLASSOPTIONcompsoc
  \section*{Acknowledgments}
\else
  \section*{Acknowledgment}
\fi
This paper is supported by the National Natural Science Foundation of China (grant 62236006), and the Key Research and Development Program of Shaanxi (No. 2022ZDLGY06-03).

\ifCLASSOPTIONcaptionsoff
  \newpage
\fi



\bibliographystyle{IEEEtran}
\bibliography{sample-base}
%

%








\end{document}